\documentclass[10pt,twocolumn,letterpaper]{article}

\usepackage{cvpr}
\usepackage{times}
\usepackage{epsfig}
\usepackage{graphicx}
\usepackage{amsmath}
\usepackage{amssymb}
\usepackage{bbm}
\usepackage{subcaption}
\usepackage{placeins}
\usepackage{color}
\usepackage{tabu}

\usepackage[pagebackref=true,breaklinks=true,letterpaper=true,colorlinks,bookmarks=false]{hyperref}
\usepackage{amsthm}

\usepackage[breaklinks=true,bookmarks=false]{hyperref}

\cvprfinalcopy 

\def\cvprPaperID{3577} 

\ifcvprfinal\fi
\begin{document}

\title{Importance Weighted Adversarial Nets for Partial Domain Adaptation}

\author{Jing Zhang, Zewei Ding, Wanqing Li, Philip Ogunbona\\
Advanced Multimedia Research Lab, University of Wollongong, Australia\\
{\tt\small jz960@uowmail.edu.au,}
{\tt\small zd027@uowmail.edu.au,}
{\tt\small wanqing@uow.edu.au,}
{\tt\small philipo@uow.edu.au}\\
}

\maketitle

\begin{abstract}
This paper proposes an importance weighted adversarial nets-based method for unsupervised domain adaptation, specific for partial domain adaptation where the target domain has less number of classes compared to the source domain. Previous domain adaptation methods generally assume the identical label spaces, such that reducing the distribution divergence leads to feasible knowledge transfer. However, such an assumption is no longer valid in a more realistic scenario that requires adaptation from a larger and more diverse source domain to a smaller target domain with less number of classes. This paper extends the adversarial nets-based domain adaptation and proposes a novel adversarial nets-based partial domain adaptation method to identify the source samples that are potentially from the outlier classes and, at the same time, reduce the shift of shared classes between domains.
\end{abstract}

\section{Introduction}
It is generally assumed that the training and test data are drawn from the same distribution in statistical learning theory. Unfortunately, this assumption does not hold in many applications. Domain adaptation~\cite{Ben-David2010, Pan2010} is a well-studied strategy to address this issue, which employs previously labeled source domain data to boost the task in a new target domain with a few or even no labeled data. Since recent advance in deep learning has shown that more transferable and domain invariant features can be extracted through deep framework, the domain adaptation techniques are also transferred from shallow learning-based~\cite{Ben-David2010, Pan2010, Long2013, Long2014, Gong2012, Fernando2013, Zhang2017} to deep learning-based~\cite{Tzeng2014, Long2015a, Long2016, Long2017, Zellinger2017, Sun2016a, Tzeng2015, Ganin2015, Tzeng2017, Bousmalis2017, Li2017, MariaCarlucci2017}. 

\begin{figure}
\includegraphics[scale=0.3]{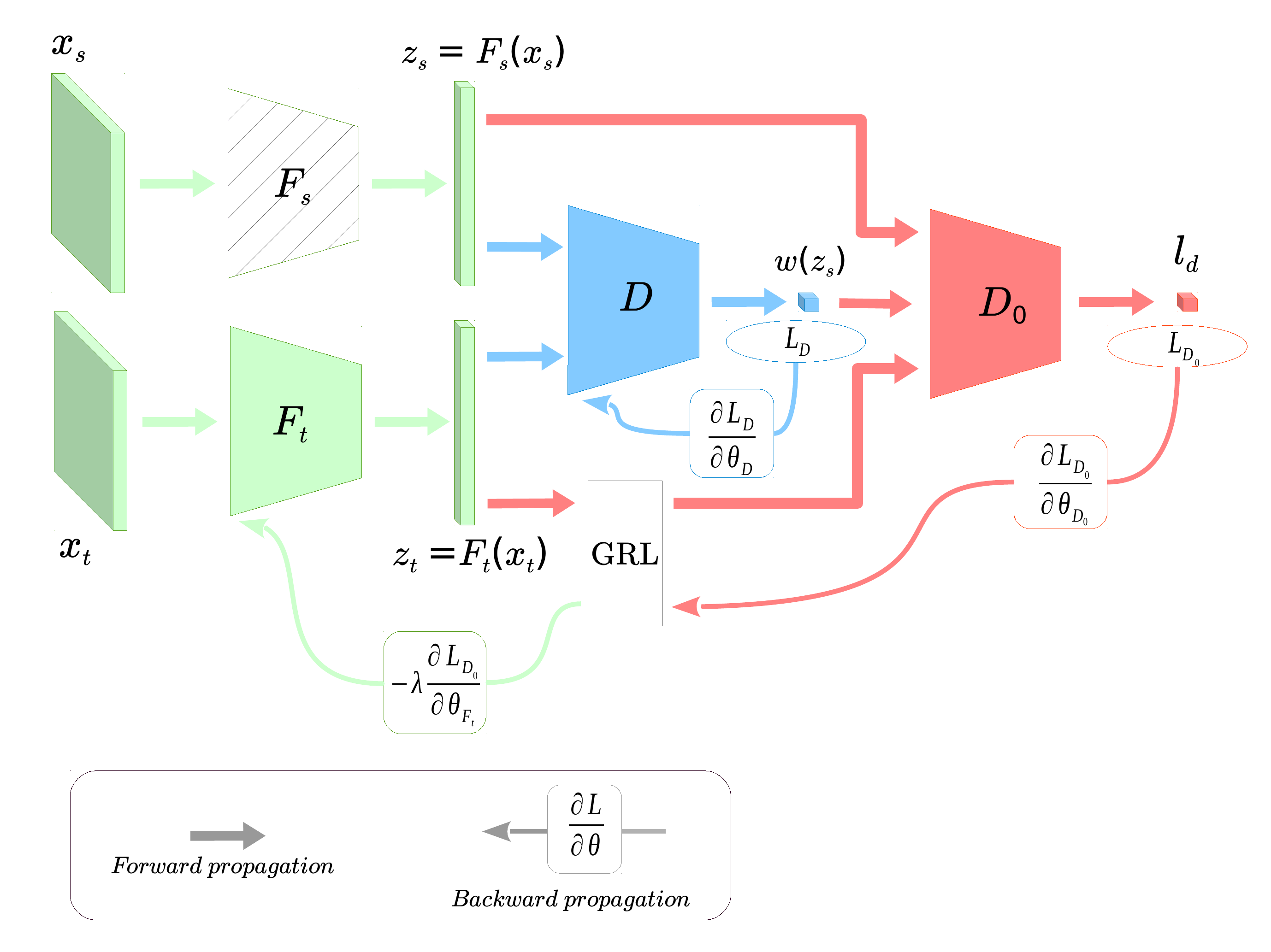}
\caption{The overview of the proposed method. The green parts are the feature extractors for source and target domains. The block filled with slashes indicates the parameters are pre-learned and will not be updated during the training procedure. The blue parts are the first domain classifier for obtaining the source sample importance weights. The red parts are the second domain classifier that plays the minimax game with the weighted source domain samples and the target samples. The GRL~\cite{Ganin2015} indicates the gradient reversal layer, which acts as an identity transformation in forward propagation while changes the sign of the gradient in backward propagation.}
\label{fig:flowgraph}
\end{figure}

The deep learning based methods have witnessed better performance compared to the shallow learning based methods. However, most of the current approaches still assume the same label spaces between the source and target domains. For example, previous deep learning-based domain adaptation methods generally follow the idea that the divergence between source and target distributions is small in the feature space and adaptation can be achieved by matching the statistic moments~\cite{Tzeng2014, Long2015a, Long2016, Long2017, Zellinger2017, Sun2016a}, or by relying on the domain adversarial nets~\cite{Tzeng2015, Ganin2015, Tzeng2017, Bousmalis2017}, or by using Batch Normalization statistics~\cite{Li2017, MariaCarlucci2017}. Since all the approaches rely on the comparison of marginal distributions between the source and target domains, the label spaces between the two domains are required to be the same for feasible adaptation. This paper is concerned with a different and more practical scenario that the target domain only has a subset of classes, referred to as partial domain adaptation (similar to~\cite{Cao2017}). In addition, there is no labeled data in the target domain and the potential number and name of the target classes are unknown. We assume that source domain is large and diverse to include all classes that appear in the target domain. 
 
Intuitively, when target domain only contains a subset of classes, it is impossible to reduce the domain shift by comparing source and target distributions directly. In other words, reducing distribution shift will not benefit the target task, since the marginal distributions between domains should not be the same essentially due to different label spaces. In this case, a natural and possible way to transfer from the source domain to the target domain is reweighting the source domain samples whose classes are likely to appear in the target domain in the distribution matching procedure. However, the target domain is unlabelled, it is not straightforward to uncover which classes are presented and which source domain samples are important for transferring. This paper proposes a weighted adversarial nets-based deep domain adaptation method for such a problem. 

An adversarial net based domain adaptation consists of a feature extractor and a domain classifier. The domain classifier aims at identifying the difference between distributions of the source and target samples to find a tighter lower bound on the true domain distribution divergence, while the feature extractor, on the other hand, reduces the distribution divergence by stepping to the opposite direction as the domain classifier. {\it This paper proposes a two domain classifier strategy} to identify the importance score of source samples.
Specifically, given any feature extractor, the output of the optimum parameters of the first domain classifier gives the probability of the sample coming from the source domain. The intuition of the weighting scheme is that if the activation of the first domain classifier is large, the sample can be almost perfectly discriminated from the target domain by the discriminator. Thus, the sample is highly likely from the outlier classes in the source domain, since the neighbourhood region of the sample covers little or no target sample at all, and a small weight is assigned to the sample. Hence, we use the activations of the first domain classifier as an indicator of the importance of each source sample to the target domain. Then the learned weights are applied to the source samples and the weighted source samples and target samples are fed into the second domain classifier for optimizing the feature extractor. {\it We have shown that the minimax two-player game between the feature extractor and the second domain classifier is theoretically equivalent to reducing the Jensen-Shannon divergence between the weighted source density and the target density.} 

The proposed methods were evaluated on three commonly used cross-domain object datasets with the setting that the target domain has a subset of classes. The results have shown that the proposed method outperforms previous domain adaptation methods to a large degree and are comparable to the state-of-the-art partial transfer method.

\section{Related Work}
The development of deep neural networks including deep convolutional neural networks (CNN)~\cite{Krizhevsky2012} has improved the visual recognition dramatically.
Recent studies have shown that deep neural networks can learn more transferable features~\cite{Bengio2012,Donahue2014,Yosinski2014}, by disentangling explanatory factors of variations underlying data samples, and grouping deep features hierarchically according to their relatedness to invariant factors. 

Recent research has shown that explicitly reducing domain divergence upon the deep learning framework can further exploit domain invariant features. Three main approaches are identified among the literature. The first is statistic moment matching based approach, i.e. maximum mean discrepancy (MMD)~\cite{Tzeng2014, Long2015a, Long2016, Long2017}, Central Moment Discrepancy (CMD)~\cite{Zellinger2017}, and second-order statistics matching~\cite{Sun2016a}. The second commonly used approach is based on an adversarial loss, which encourages samples from different domains to be non-discriminative with respect to domain labels, i.e. domain adversarial nets-based adaptation methods~\cite{Tzeng2015, Ganin2015, Tzeng2017, Bousmalis2017} borrowing the idea of GAN~\cite{Goodfellow2014}. The third approach uses Batch Normalization statistics~\cite{Li2017, MariaCarlucci2017}, which aligns the source and target distributions to a canonical one. However, all of these approaches rely on the marginal distribution matching in the feature space and thus the label spaces between domains are assumed to be identical for feasible adaptation.

The method proposed by Ganin \etal~\cite{Ganin2015} is related to our work. They use a single domain classifier to regularize the extracted features to be indiscriminate with respect to the different domains. However, they assume the existence of a shared feature space between domains where the distribution divergence is small. By contrast, we use two different feature extractors for respective domains to learn more domain specific features. In addition, we weight the source domain samples when learning the two domain classifiers, such that the outlier samples from the source domain will be ignored for more effective transfer, especially when the target domain only contains a subset of classes of the source domain. Another related work is~\cite{Tzeng2017}, which also learns two different feature extractors by unsharing the weights in the adversarial nets-based framework. However, it assumes the identical label space between domains and cannot deal with the partial domain adaptation as addressed by this paper. 

A recent report by Cao \etal~\cite{Cao2017} also addresses the problem of transferring from big source domain to the target domain with a subset of classes. SAN trains a separate domain classifier for each class and introduces both instance-level and class-level weights according to the class probabilities given by label predictor.
There are fundamental differences between the proposed method and the methods in~\cite{Cao2017}. Firstly, their method uses a shared feature extractor for both domains. Secondly, our method only requires two domain classifiers rather than multiple domain classifiers (one per source class) which makes their method hardly scalable to a source data with a large number of classes and is computationally expensive. Lastly, our method does not require class level weight and hence be able to deal with imbalanced target data because if a class level weight is applied, the target classes with a smaller number of samples may not be able to be classified well after adaptation. 

\section{Proposed Method}
This section presents the proposed method in details. It begins with the definitions of terminologies. The source domain data denoted as $X_s\in{\mathbb{R}^{D\times n_s}}$ are draw from distribution $p_s(\mathbf{x})$ and the target domain data denoted as $X_t\in{\mathbb{R}^{D\times n_t}}$ are draw from distribution $p_t(\mathbf{x})$, where D is the dimension of the data instance, $n_s$ and $n_t$ are number of samples in source and target domain respectively. We focus on the unsupervised domain adaptation problem which assumes that there are sufficient labeled source domain data, $\mathcal{D}_s = \{(\mathbf{x}_i^s,y_i^s)\}_{i=1}^{n_s}$, $\mathbf{x}_i^s\in{\mathbb{R}^D}$, and unlabeled target domain data, $\mathcal{D}_t = \{(\mathbf{x}_j^t)\}_{j=1}^{n_t}$, $\mathbf{x}_j^t\in{\mathbb{R}^D}$, in the training stage. The feature spaces are assumed same: $\mathcal{X}_s = \mathcal{X}_t$ while the target domain label space is contained in the source domain label space $\mathcal{Y}_t \subseteq \mathcal{Y}_s $. In addition, due to the domain shift, $p_s(\mathbf{x})\neq{p_t(\mathbf{x})}$ even when the label spaces between domains are the same. 
\subsection{Adversarial Nets-based Domain Adaptation}
The works in~\cite{Ganin2015,Tzeng2017} apply a domain classifier on the general feed-forward models to form the adversarial nets-based domain adaptation methods. The general idea is to learn both class discriminative and domain invariant features, where the loss of the label predictor of the source data is minimized while the loss of the domain classifier is maximized. Specifically, the adversarial nets-based domain adaptation framework is similar to the original GAN with minimax loss:
\begin{equation}
\begin{split}
\min_{F_s,F_t}\max_{D}\mathcal{L}(D, F_s, F_t)&=\mathbb{E}_{\mathbf{x}\sim p_s(\mathbf{x})}[\log D(F_s(\mathbf{x}))]\\
&+\mathbb{E}_{\mathbf{x}\sim p_t(\mathbf{\mathbf{x}})}[\log (1-D(F_t(\mathbf{x})))]
\end{split}
\label{eqt:DANN}
\end{equation}
where $F_s$ and $F_t$ are the feature extractors for source and target data respectively, which can be identical~\cite{Ganin2015} (shared weights) or distinct~\cite{Tzeng2017} (unshared weights), and $D$ is the domain classifier. The $D$ is a binary domain classifier (corresponding to the discriminator in original GAN) with all the source data labelled as 1 and all the target data labelled as 0. Maximizing the minimax loss with respect to the parameters of $D$ yields a tighter lower bound on the true domain distribution divergence, while minimizing the minimax loss with respect to the parameters of $F$ minimizes the distribution divergence in the feature space.

In this paper, we adopt the unshared feature extractors for source and target domains to capture more domain specific features than a shared feature extractor as reported in~\cite{Zhang2017, Tzeng2017} and to train the source discriminative model separately. We follow a similar procedure as~\cite{Tzeng2017} to train the source discriminative model $C(F_s(\mathbf{x}))$ for classification task by learning the parameters of the source feature extractor $F_s(\mathbf{x})$ and classifier $C$:
\begin{equation}
\min_{F_s,C}\mathcal{L}_{s} = \mathbb{E}_{\mathbf{x},y \sim p_s(\mathbf{x}, y)} L(C(F_s(\mathbf{x})), y)
\label{eqt:lossS}
\end{equation}
where $L$ is the empirical loss for source domain classification task and the cross entropy loss is used in this paper. 

Given the learned $F_s(\mathbf{x})$, a domain adversarial loss is used to reduce the shift between domains by optimizing $F_t(\mathbf{x})$ and $D$:
\begin{equation}
\begin{split}
\min_{F_t}\max_{D}\mathcal{L}(D, F_s, F_t)&=\mathbb{E}_{\mathbf{x}\sim p_s(\mathbf{x})}[\log D(F_s(\mathbf{x}))]\\
&+\mathbb{E}_{\mathbf{x}\sim p_t(\mathbf{\mathbf{x}})}[\log (1-D(F_t(\mathbf{x})))]
\end{split}
\label{eqt:DANNFt}
\end{equation}
To avoid a degenerate solution, we initialize $F_t$ using the parameter of $F_s$ by following~\cite{Tzeng2017}.

Given $F_s(\mathbf{x})$ (corresponding to real images in GAN), for any $F_t(\mathbf{x})$ (corresponding to generated images in GAN), the optimum $D$ is obtained at:
\begin{equation}
D^{*}(\mathbf{z})=\frac{p_s(\mathbf{z})}{p_s(\mathbf{z})+p_t(\mathbf{z})}
\label{eqt:DANNmax}
\end{equation}
where $\mathbf{z}=F(\mathbf{x})$ is the sample in the feature space after feature extraction networks. Similar to~\cite{Goodfellow2014}, we give the proof of Equation~\ref{eqt:DANNmax} as follows.
\begin{proof}
For any $F_s(\mathbf{x})$ and $F_t(\mathbf{x})$, the training criterion for the domain classifier $D$ is to maximize Equation~\ref{eqt:DANNFt}:
\begin{equation}
\begin{split}
\max_{D}\mathcal{L}(D, F_s, F_t)=&\int_{\mathbf{x}}p_s(\mathbf{x})\log D(F_s(\mathbf{x}))\\
&+ p_t(\mathbf{\mathbf{x}})\log (1-D(F_t(\mathbf{x})))d\mathbf{x}\\
=&\int_{\mathbf{z}}p_s(\mathbf{z})\log D(\mathbf{z})\\
&+ p_t(\mathbf{z})\log (1-D(\mathbf{z}))d\mathbf{z}
\end{split}
\label{eqt:DANNmaxD}
\end{equation}
We take the partial differential of the objective~\ref{eqt:DANNmaxD} with respect to $D$, $\frac{\partial\mathcal{L}(D, F)}{\partial D}$, and achieves its maximum in [0, 1] at~\ref{eqt:DANNmax}, where the Leibniz’s rule is used to exchange the order of differentiation and integration.
\end{proof}
\subsection{Importance Weighted Adversarial Nets-based Domain Adaptation}
\paragraph{Sample weights learning} In the minimax game of Equation~\ref{eqt:DANNFt}, the domain classifier is given by
\begin{equation}
D(\mathbf{z})=p(y = 1 | \mathbf{z})= \sigma (a(\mathbf{z}))
\end{equation}
where $\sigma$ is the logistic sigmoid function. Suppose that the domain classifier has converged to its optimal value for the current feature extractor, the output value of the domain classifier gives the likelihood of the sample coming from source distribution. Thus, if the $D^{*}(\mathbf{z})\approx 1$, then the sample is highly likely come from the outlier classes in the source domain, since the region that covers the sample has little or no target sample at all and can be almost perfectly discriminated from target distribution by the domain classifier. The contribution of these samples should be small such that both the domain classifier and feature extractor will ignore them. On the other hand, if $D^{*}(\mathbf{z})$ is small, the sample is more likely from the shared classes between domains. These samples should be given a larger importance weight to reduce the domain shift on the shared classes. Hence, the weight function should be inversely related to $D^{*}(\mathbf{z})$ and a natural way to define the importance weights function of the source samples is:
\begin{equation}
\tilde{w}(\mathbf{z})= 1-D^{*}(\mathbf{z})=\frac{1}{\frac{p_s(\mathbf{z})}{p_t(\mathbf{z})}+1} 
\end{equation}
It can be seen that if $D^{*}(\mathbf{z})$ is large, $\tilde{w}(\mathbf{z})$ is small and thus $\frac{p_s(\mathbf{z})}{p_t(\mathbf{z})}$ is large. Hence,  the weights for source samples from outlier classes will be smaller than the shared class samples. Note that the weights function is also a function of density ratio between source and target features, which further verifies the reasonableness of the weights function, since the neighbourhood region of the sample that covers little or no target sample will be assigned a small weight. Our purpose is to obtain the relative importance of source samples, suggesting that the samples from outlier classes should be assigned a relatively small weight than the samples from the shared classes. Hence, the weights are normalized as follows
\begin{equation}
w(\mathbf{z})= \frac{\tilde{w}(\mathbf{z})} {\mathbb{E}_{\mathbf{z}\sim p_s(\mathbf{z})}\tilde{w}(\mathbf{z})}
\end{equation} 
such that $\mathbb{E}_{\mathbf{z}\sim p_s(\mathbf{z})} w(\mathbf{z}_i)=1$. Note that the weights are defined as a function of the domain classifier. Thus if we apply the weights on the same domain classifier, the theoretical results of the minimax game will not be reducing the Jensen-Shannon divergence between two densities (since the optimum discriminator (e.g. Equation~\ref{eqt:DANNmax}) will not be the ratio between the source density and the sum of the source and target densities due to the introducing of the weight function which is also a function of $D$). Hence, we propose to solve this issue by applying the second domain classifier on the extracted features, namely $D_0$, for comparing the weighted source data and the target data. In this way, the first domain classifier $D$ is only used for obtaining the importance weights for the source domain based on $F_s$ and the current $F_t$. Thus, the gradient of $D$ will not be back-propagated for updating $F_t$, since the gradients of $D$ are learned on unweighted source samples and would not be a good indicator for reducing domain shift on the shared classes. After all, it is $D_0$ (with the weighted source data and the target data) who plays the minimax game with $F_t$ to reduce the shift on the shared classes.

After adding the importance weights to the source samples for the domain classifier $D_0$, the objective function of weighted domain adversarial nets $\mathcal{L}_{w}(D_0, F)$ is:
\begin{equation}
\begin{split}
\min_{F_t}\max_{D_0}\mathcal{L}_{w}(D_0, F_s, F_t)&=\mathbb{E}_{\mathbf{x}\sim p_s(\mathbf{x})}[w(\mathbf{z})\log D_0(F_s(\mathbf{x}))]\\
&+\mathbb{E}_{\mathbf{x}\sim p_t(\mathbf{\mathbf{x}})}[\log (1-D_0(F_t(\mathbf{x})))]
\end{split}
\label{eqt:weightedDANN}
\end{equation}
where the $w(\mathbf{z})$, as a function of $D$, is independent of $D_0$ and can be seen as a constant. Thus, given $F_s$ and $D$, for any $F_t$, the optimum $D_0$ of the weighted adversarial nets is obtained at:
\begin{equation}
D_0^{*}(\mathbf{z})=\frac{w(\mathbf{z})p_s(\mathbf{z})}{w(\mathbf{z})p_s(\mathbf{z})+p_t(\mathbf{z})}
\label{eqt:weightedDANNmax}
\end{equation}
Note that since we normalized the importance weights $w(\mathbf{z})$, the $w(\mathbf{z})p_s(\mathbf{z})$ is still a probability density function:
\begin{equation}
\mathbb{E}_{\mathbf{z}\sim p_s(\mathbf{z})} w(\mathbf{z}_i) = \int w(\mathbf{z})p_s(\mathbf{z}) d\mathbf{z} = 1
\end{equation}

Given the optimum $D_0$, the minimax game of~\ref{eqt:weightedDANN} can be reformulated as:
\begin{equation}
\begin{split}
\mathcal{L}_w(F_t)=&\mathbb{E}_{\mathbf{x}\sim p_s(\mathbf{x})}[w(\mathbf{z})\log D_0^{*}(F_s(\mathbf{x}))]\\
&+\mathbb{E}_{\mathbf{x}\sim p_t(\mathbf{\mathbf{x}})}[\log (1-D_0^{*}(F_t(\mathbf{x})))]\\
=&\int_{\mathbf{z}}w(\mathbf{z})p_s(\mathbf{z})\log \frac{w(\mathbf{z})p_s(\mathbf{z})}{w(\mathbf{z})p_s(\mathbf{z})+p_t(\mathbf{z})}\\
&+ p_t(\mathbf{\mathbf{z}})\log \frac{p_t(\mathbf{z})}{w(\mathbf{z})p_s(\mathbf{z})+p_t(\mathbf{z})}d\mathbf{z}\\
=&-\log(4)+2\cdot JS(w(\mathbf{z})p_s(\mathbf{z})\|p_t(\mathbf{z}))
\end{split}
\label{eqt:weightedDANNmin}
\end{equation}
Hence, the weighted adversarial nets-based domain adaptation is essentially reducing the Jensen-Shannon divergence between the weighted source density and the target density in the feature space, which obtains it's optimum on $w(\mathbf{z})p_s(\mathbf{z})=p_t(\mathbf{z})$.
\vspace{-1em}
\paragraph{Target data structure preservation} Since the target domain does not have labels, it is important to preserve the data structure for effective transfer. If the shift between the weighted source distribution and target distribution in the feature space is small, the classifier $C$ learned from the source data can be directly used for the target domain. Here, we further constrain $F_t$ by employing the entropy minimization principle~\cite{Grandvalet2005} to encourage the low-density separation between classes:
\begin{equation}
\min_{F_t} \mathbb{E}_{\mathbf{x}\sim p_t(\mathbf{x})}H(C(F_t(\mathbf{x})))
\label{eqt:lossT}
\end{equation}
where $H(\cdot)$ is the information entropy function.
Since the source classifier C is directly applied to the adapted target features, the target entropy minimization is only used to constrain $F_t$, which is different from previous usage~\cite{Long2016, Cao2017}. We argue that if target entropy minimization is applied on both feature extractor and classifier as in~\cite{Long2016, Cao2017}, a side effect is that the target samples may easily be stuck into a wrong class due to the large domain shift in the early stage of training and hard to be corrected later on. By contrast, if target entropy minimization is only used to constrain $F_t$, it will reduce the side effect.
\vspace{-1em}
\paragraph{Overall objective function} Hence, the overall objectives of the weighted adversarial nets-based method are:
\begin{equation}
\begin{split}
&\min_{F_s,C}\mathcal{L}_{s}(F_s, C) = -\mathbb{E}_{\mathbf{x}, y\sim p_s(\mathbf{x}, y)} \sum_{k=1}^{K}\mathbbm{1}_{[k=y]}\log C(F_s(\mathbf{x})) \\
&
\begin{split}
\min_{D}\mathcal{L}_D(D, F_s, F_t) = &-\big(\mathbb{E}_{\mathbf{x}\sim p_s(\mathbf{x})}[\log D(F_s(\mathbf{x}))]\\
&+\mathbb{E}_{\mathbf{x}\sim p_t(\mathbf{\mathbf{x}})}[\log (1-D(F_t(\mathbf{x})))]\big)
\end{split}\\
&
\begin{split}
\min_{F_t}\max_{D_0}&\mathcal{L}_{w}(C, D_0, F_s, F_t)=\\
&\gamma \mathbb{E}_{\mathbf{x}\sim p_t(\mathbf{x})} H(C(F_t(\mathbf{x})))\\
&+ \lambda \big( \mathbb{E}_{\mathbf{x}\sim p_s(\mathbf{x})}[w(\mathbf{z})\log D_0(F_s(\mathbf{x}))]\\
&+\mathbb{E}_{\mathbf{x}\sim p_t(\mathbf{\mathbf{x}})}[\log (1-D_0(F_t(\mathbf{x})))] \big)
\end{split}
\end{split}
\label{eqt:overall}
\end{equation}
where $\lambda$ is the tradeoff parameter.
The objectives are optimized in stages. $F_s$ and $C$ are pre-trained on the source domain data and fixed afterwards. Then the $D$, $D_0$ and $F_t$ are optimized simultaneously without the need of revisiting $F_s$ and $C$. Note that $D$ is only used for obtaining the importance weights for the source domain using $F_s$ and current $F_t$, while $D_0$ plays the minimax game with the target domain feature extractor for updating $F_t$. To solve the minimax game between $F_t$ and $D_0$, we can either iteratively train the two objectives respectively similar to GAN, or insert a gradient reversal layer (GRL)~\cite{Ganin2015} to multiply the gradient by -1 for the feature extractor to learn the feature extractor and domain classifier simultaneously. In this paper, we choose to use the GRL for solving the problem for the fair comparison with previous methods. The proposed architecture can be found in Figure~\ref{fig:flowgraph}.

\section{Experiments}
\subsection{Set-ups}
\begin{figure}[ht!]
\begin{tabular}{cccc}
\begin{minipage}{.09\textwidth}
\includegraphics[scale=0.1]{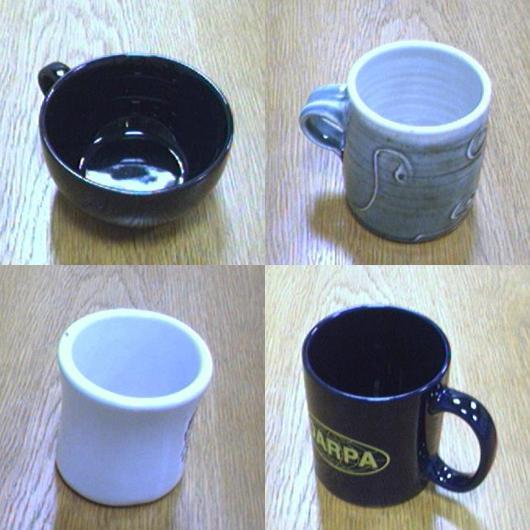}
\end{minipage} & 
\begin{minipage}{.09\textwidth}
\includegraphics[scale=0.1]{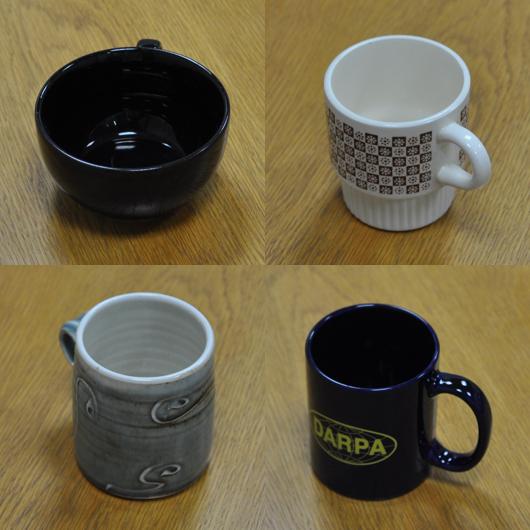}
\end{minipage} &
\begin{minipage}{.09\textwidth}
\includegraphics[scale=0.1]{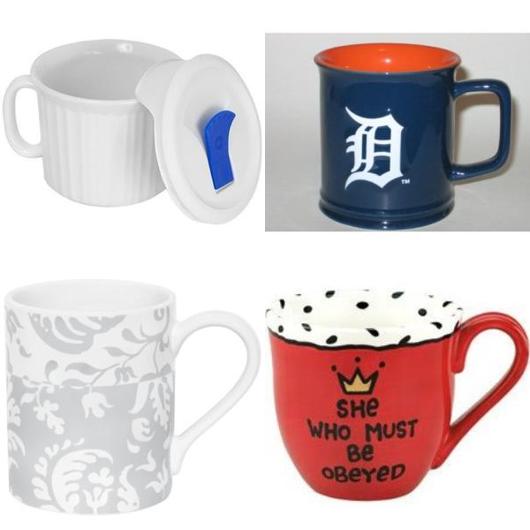}
\end{minipage} & 
\begin{minipage}{.09\textwidth}
\includegraphics[scale=0.1]{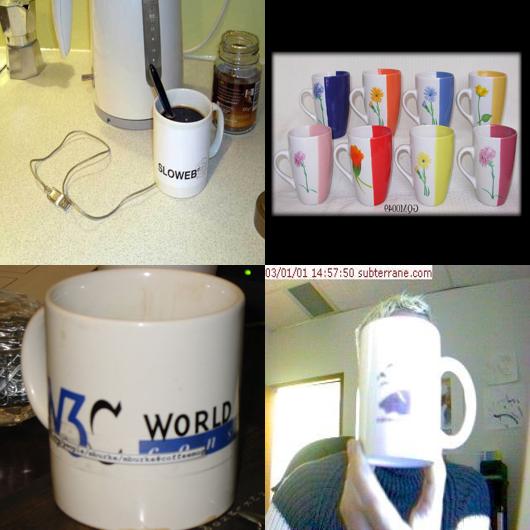}
\end{minipage} \\
Webcam & DSLR & Amazon & Caltech\\
\end{tabular}
\vspace{-1em}
\caption{Sample images of Caltech and Office datasets.}
\label{fig:officeDatasets}
\end{figure}

\paragraph{Datasets} The proposed method is evaluated on three commonly used real world cross-domain object recognition datasets. The public Office+Caltech-10 object datasets released by Gong et al.~\cite{Gong2012} contains four different domains: Amazon (images downloaded from online merchants), Webcam (low-resolution images by a web camera), DSLR (high-resolution images by a digital SLR camera), and Caltech-256~\cite{Griffin2007}, where the first three domains come from Office-31~\cite{Saenko2010}. Ten shared classes of the four domains form the Office+Caltech-10 dataset. Figure~\ref{fig:officeDatasets} shows the sample images of the four different domains. When a domain is used as the target domain, the first five classes are selected. We denote the source domains with 10 classes as A10, W10, D10, and C10, while the target domains with 5 classes are denoted as A5, W5, D5, and C5.

We also evaluate our method on the Office-31 dataset studied by Saenko \etal~\cite{Saenko2010}, which consists of three different domains: Amazon, DSLR, and Webcam. Compared to Office+Caltech-10, more classes (31 classes) are involved. We follow the experimental setting of ~\cite{Cao2017} to transfer from one domain with the 31 categories to another domain with 10 categories (which are the shared classes between Office31 and Caltech-256~\cite{Griffin2007}). Hence, the three source domains are denoted as A31, W31, and D31, and the three target domains are denoted as A10, W10, and D10.

To evaluate on the larger scale datasets, we conducted the experiments on three pairs of domains formed by Caltech256$\rightarrow$Office10 datasets, where the source domain is Caltech-256 dataset with 256 classes and the target domains are three Office domains with 10 shared classes (denoted as Office-10) between Caltech-256 and Office-31.

\vspace{-1em}
\paragraph{Baselines and Settings} 
The proposed method is compared with the baseline that finetuning the CNN using source data only (AlexNet+bottleneck) and several state-of-the-art deep learning-base domain adaptation methods: RevGrad~\cite{Ganin2015}, RTN~\cite{Long2016}, ADDA-grl~\cite{Tzeng2017}, and SAN~\cite{Cao2017}. Note that ADDA-grl is a variant of the original ADDA~\cite{Tzeng2017} method, where the minimax game is not trained iteratively but using the GRL layer as done in our method for fair comparison. Thus the ADDA-grl can be seen as a special case of our method without the weighting scheme.

Since the cross-domain datasets are relatively small, to successfully train a deep network, we finetune the AlexNet pre-trained on ImageNet similar to previous deep learning-based domain adaptation methods~\cite{Ganin2015, Long2016}. For the fair comparison, the same network architectures as the RevGrad method~\cite{Ganin2015} are used for feature extractors and domain classifiers. Specifically, the feature extractors are the AlexNet without $fc8$ layer, and an additional bottleneck layer is added to $fc7$ layer with the dimension of 256. The two domain classifiers are with the same architecture, which are three fully connected layers (1024$\rightarrow$1024$\rightarrow$1) attached to the bottleneck layer. The $F_s$ is obtained from the source domain data by finetuning the AlexNet+bottleneck.

To avoid the noisy signal at the early stage of training procedure, we use similar schedule method as~\cite{Ganin2015} for the tradeoff parameter to update $F_t$ by initializing it at 0 and gradually increasing to a pre-defined upper bound. The schedule is defined as: $\lambda = \frac{2\cdot u}{1+\exp (-\alpha \cdot p)}-u$,
where $p$ is the training progress linearly changing from 0 to 1, $\alpha=1$, and $u$ is the upper bound set to 0.1 in our experiments.

\subsection{Results and Analysis}
\paragraph{Evaluation of partial domain adaptation}
{Table~\ref{tab:officeCal}, Table~\ref{tab:office} and Table~\ref{tab:cal}} show the results of the proposed methods compared with the baseline methods, where the results of SAN methods are directly copied from the original paper~\cite{Cao2017}. The proposed ($\gamma=0$) in Table~\ref{tab:officeCal} and Table~\ref{tab:office} is the variant of the proposed method without the target domain entropy minimization term. The results show that the proposed methods outperform AlexNet+bottleneck, RevGrad,  RTN, and ADDA-grl to a large degree, and also comparable to the state-of-the-art partial domain adaptation method SAN on most of the datasets. 

\begin{figure*}[!ht]
    \centering
    \begin{subfigure}[t]{0.32\textwidth}
        \centering
        \includegraphics[scale=0.19]{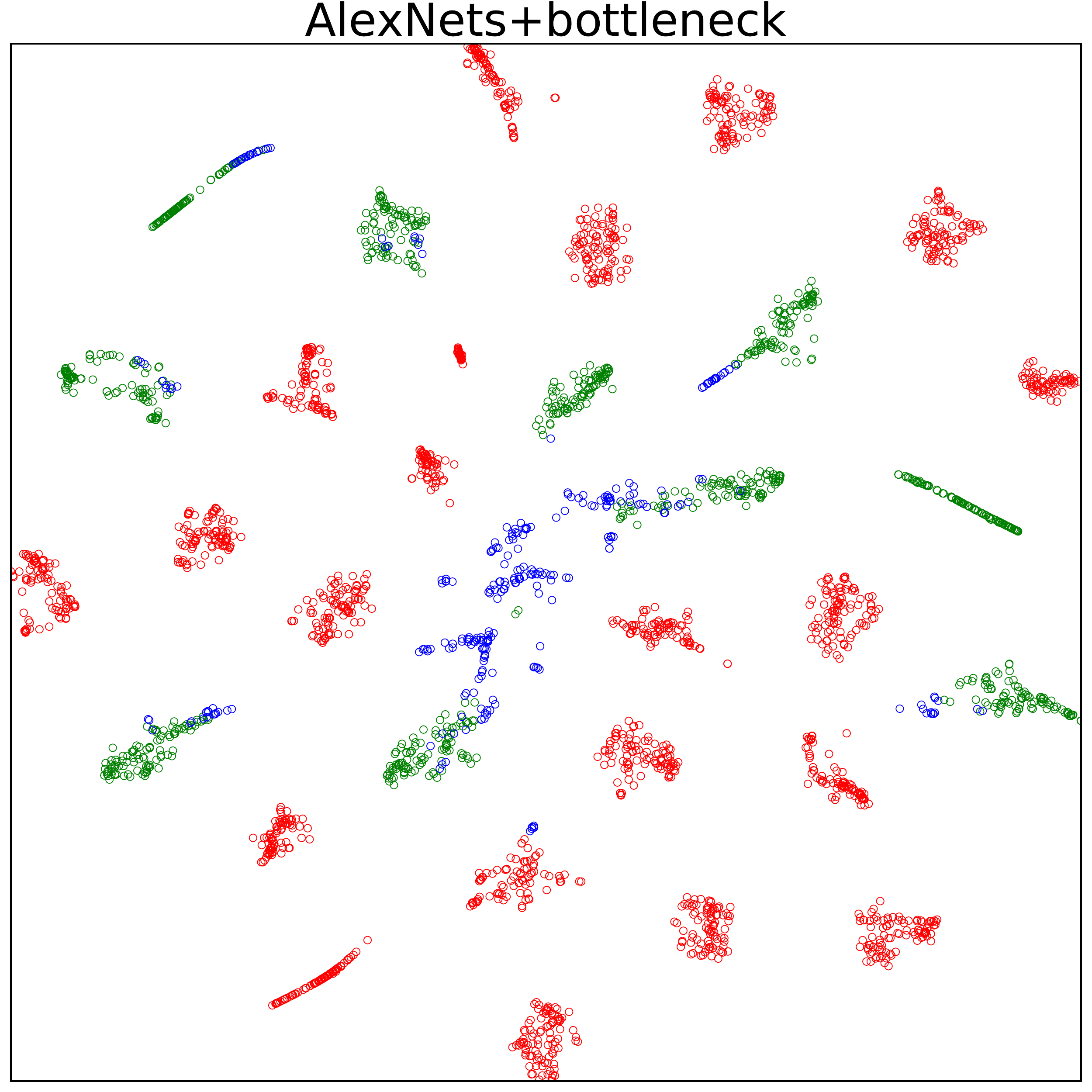}
\caption{}
\label{fig:DA1}
    \end{subfigure}%
~
	\begin{subfigure}[t]{0.32\textwidth}
        \centering
        \includegraphics[scale=0.19]{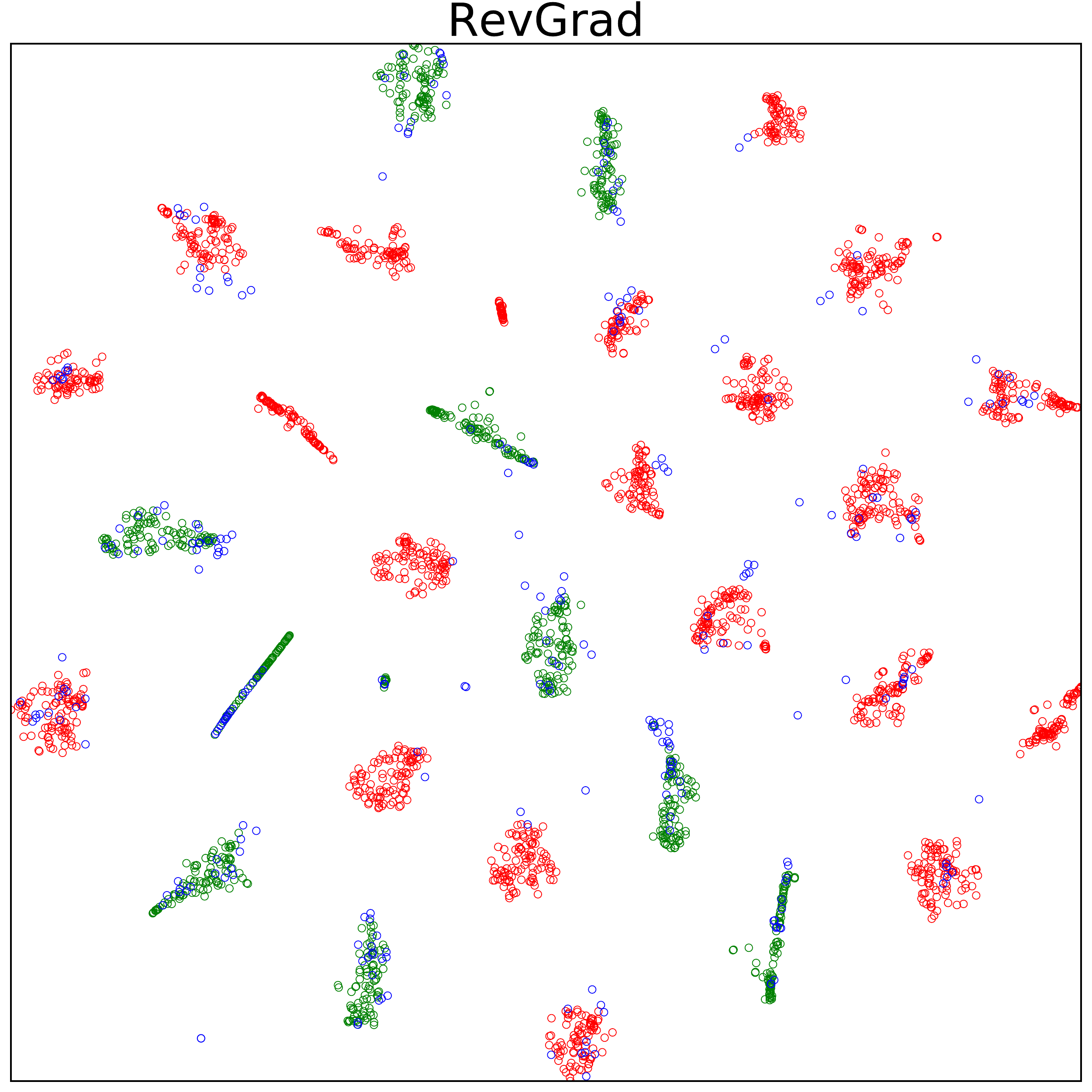}
\caption{}
\label{fig:DA2}
    \end{subfigure}
~
\begin{subfigure}[t]{0.32\textwidth}
        \centering
        \includegraphics[scale=0.19]{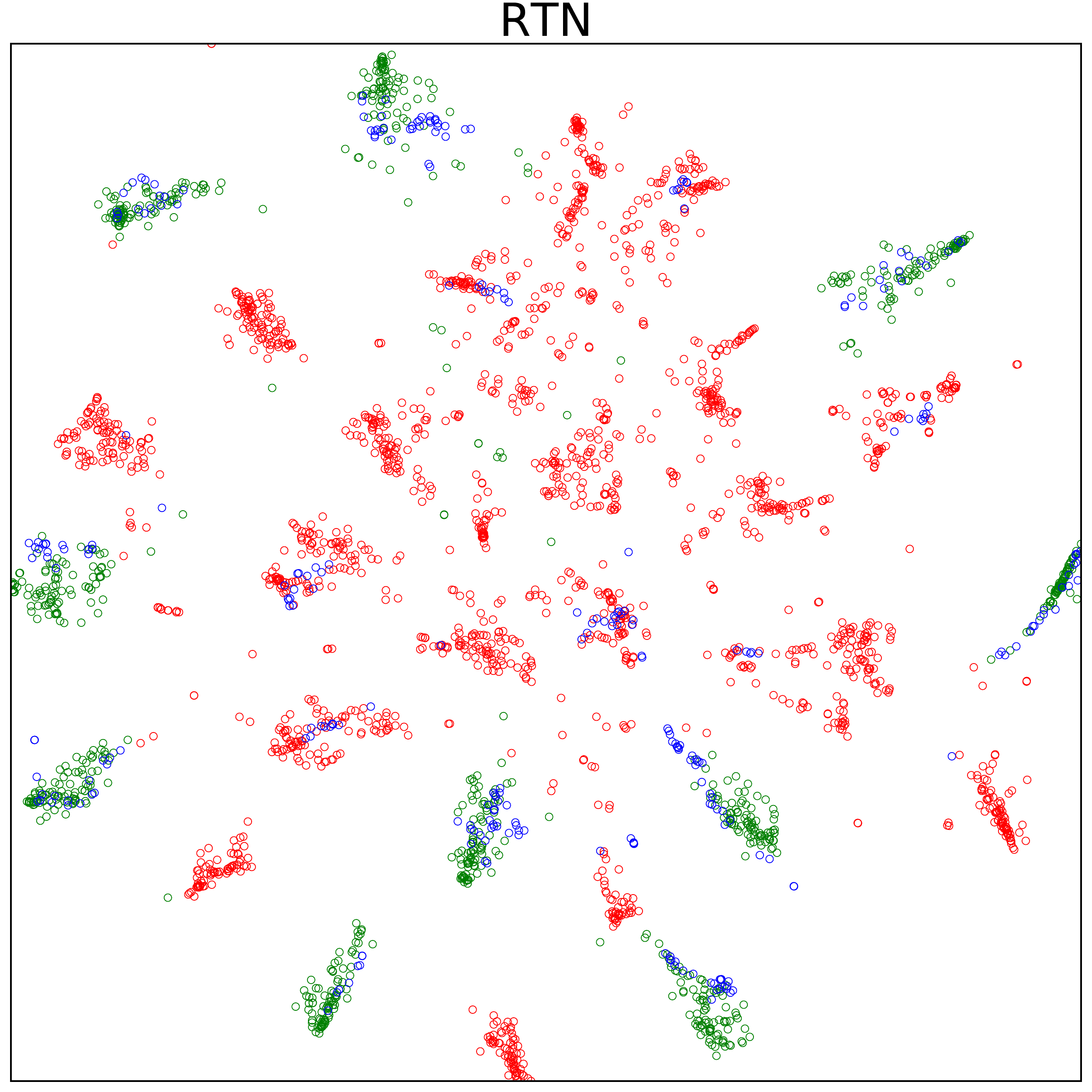}
\caption{}
\label{fig:DA3}
    \end{subfigure}\\
\begin{subfigure}[t]{0.4\textwidth}
        \centering
        \includegraphics[scale=0.19]{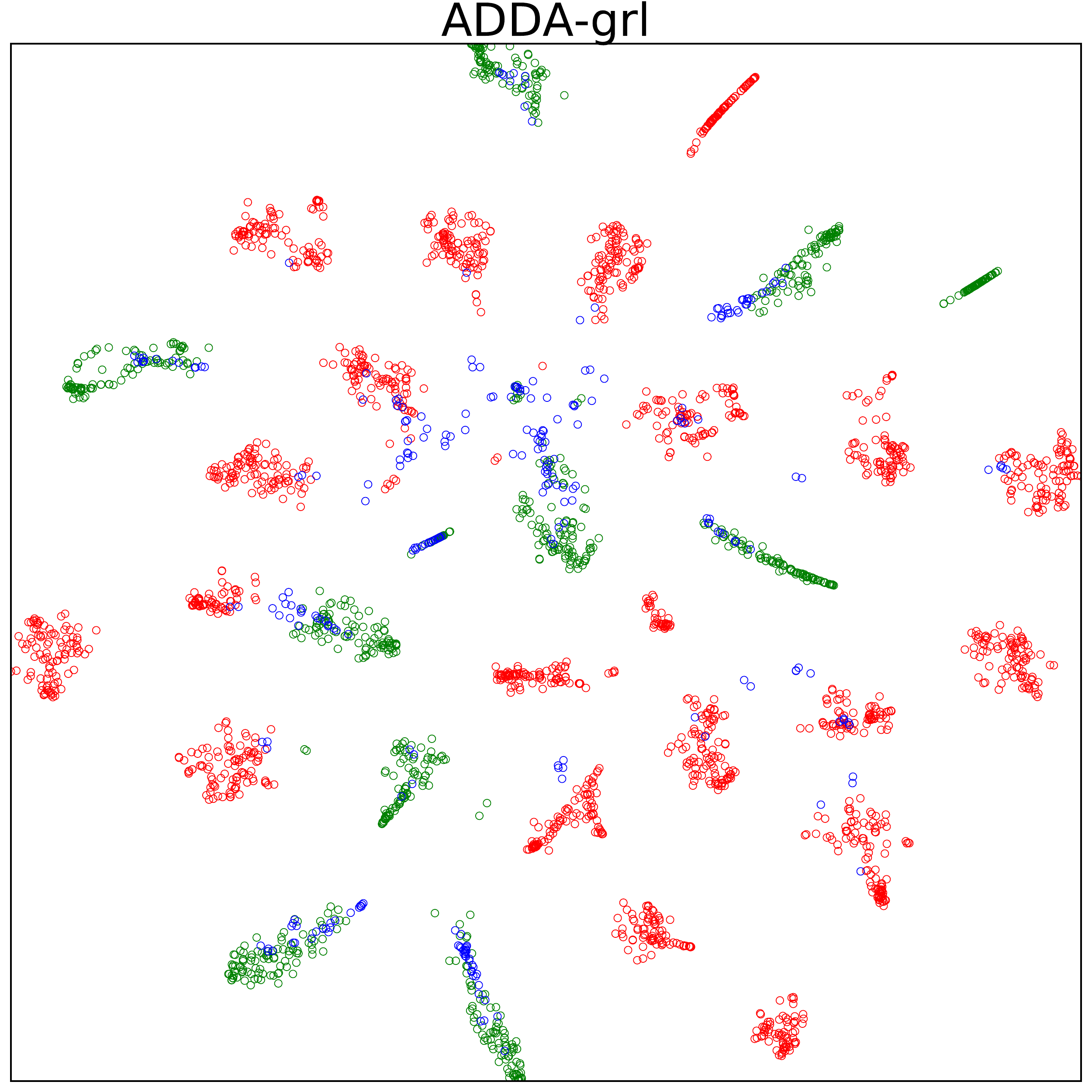}
\caption{}
\label{fig:DA4}
    \end{subfigure}
~
\begin{subfigure}[t]{0.4\textwidth}
        \centering
        \includegraphics[scale=0.19]{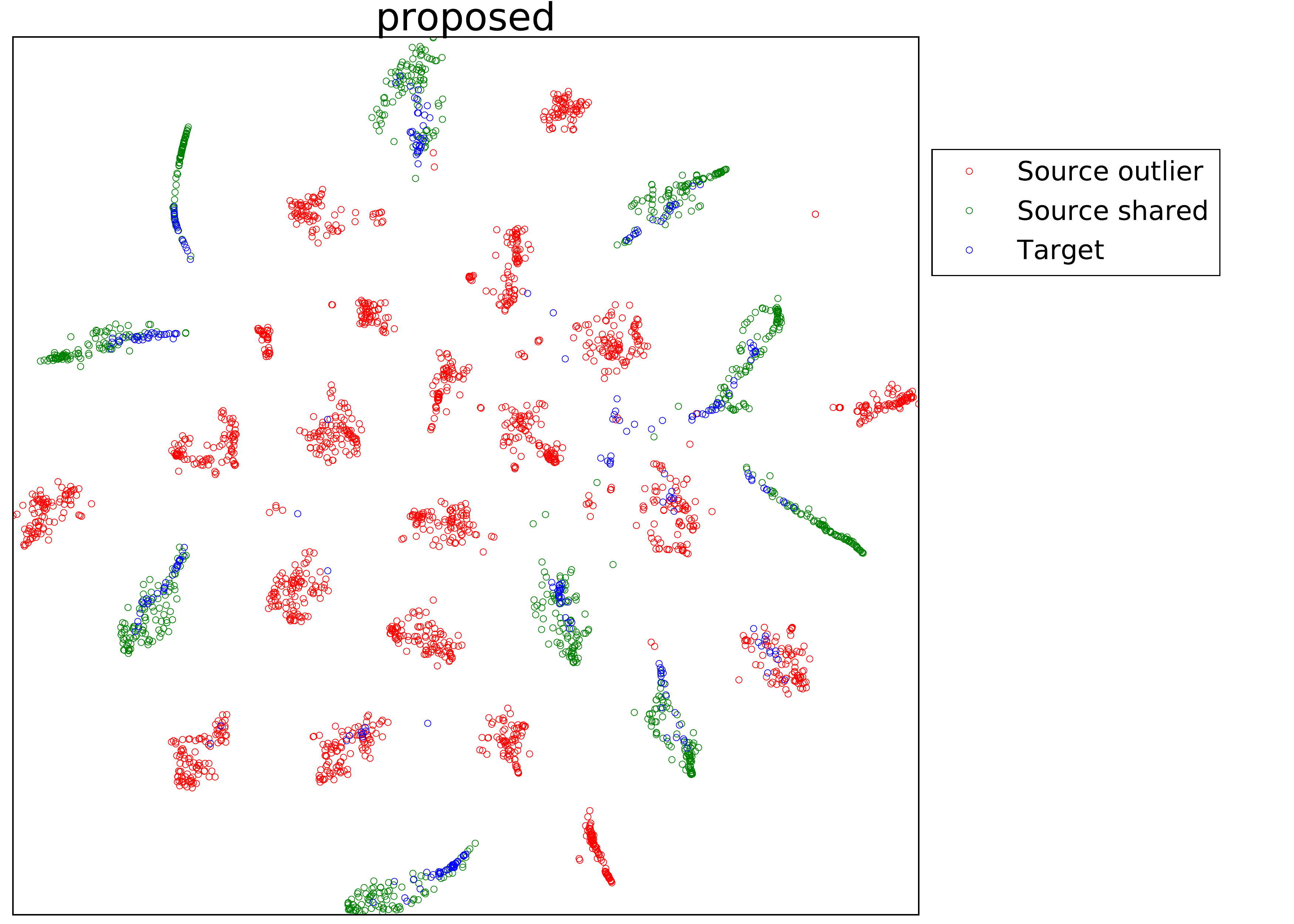}
\caption{}
\label{fig:DA5}
    \end{subfigure}
    \vspace{-1em}
    \caption{The t-SNE visualization of the activations of baseline methods and the proposed method. The blue dots are expected to be aligned with green dots for effective domain adaptation.}
    \label{fig:bottleneck}
\end{figure*}

We also illustrate the A31$\rightarrow$W10 data activations of the bottleneck layer for AlexNet+bottleneck, RevGrad, RTN, ADDA-grl, and the proposed method in Figure~\ref{fig:bottleneck}, where the red dots (outlier classes) and green dots (shared classes) indicate the source domain samples while the blue dots represent the target samples. The alignment is effective if the blue dots are well aligned with green dots. It shows that our method can effectively match the target classes into the relevant source domain classes compared to the baseline methods.

The RevGrad is an adversarial nets-based method with the domain classifier as a regularization for the source domain classification task. Since the adversarial training procedure only reduces the marginal distributions between domains without considering the conditional distributions, the RevGrad method obtains even much poorer results than the AlexNet+bottleneck baseline on most of the domain pairs in both datasets. Figure~\ref{fig:DA2} also verifies that though the target domain only contains ten classes, the samples will spread to all the 31 classes in the source domain. Instead of using the adversarial loss, the RTN method reduces the domain shift based on MMD criterion. In addition, the unshared classifiers for source and target domains are proposed using a residual block and the target domain entropy minimization is applied for preserving the target domain structure. Figure~\ref{fig:DA3} shows that the target samples are not spread to all the classes as in RevGrad due to the target domain structure preservation term. However, the RTN still performs unsatisfied for target domain classification task and the negative transfer can also be seen.
Thus, though the residual nets and the target entropy minimization are involved the source domain outlier classes that do not appear in the target domain can still ruin the performance.

The ADDA-grl can be seen as the unweighted version of our method. For the fair comparison, we use exactly the same sets of parameters for ADDA-grl and our method. The results show that the proposed method outperforms ADDA-grl on most of the domain pairs. Thus the proposed weighting scheme can effectively detect the outlier classes and reduce the shift between the shared classes. Figure~\ref{fig:DA4} and Figure~\ref{fig:DA5} compares the activations of the two methods. The target data in the proposed method is better aligned with the selected source classes than in ADDA-grl.

\begin{figure*}[!h]
\minipage{0.47\textwidth}
  \includegraphics[scale=0.31]{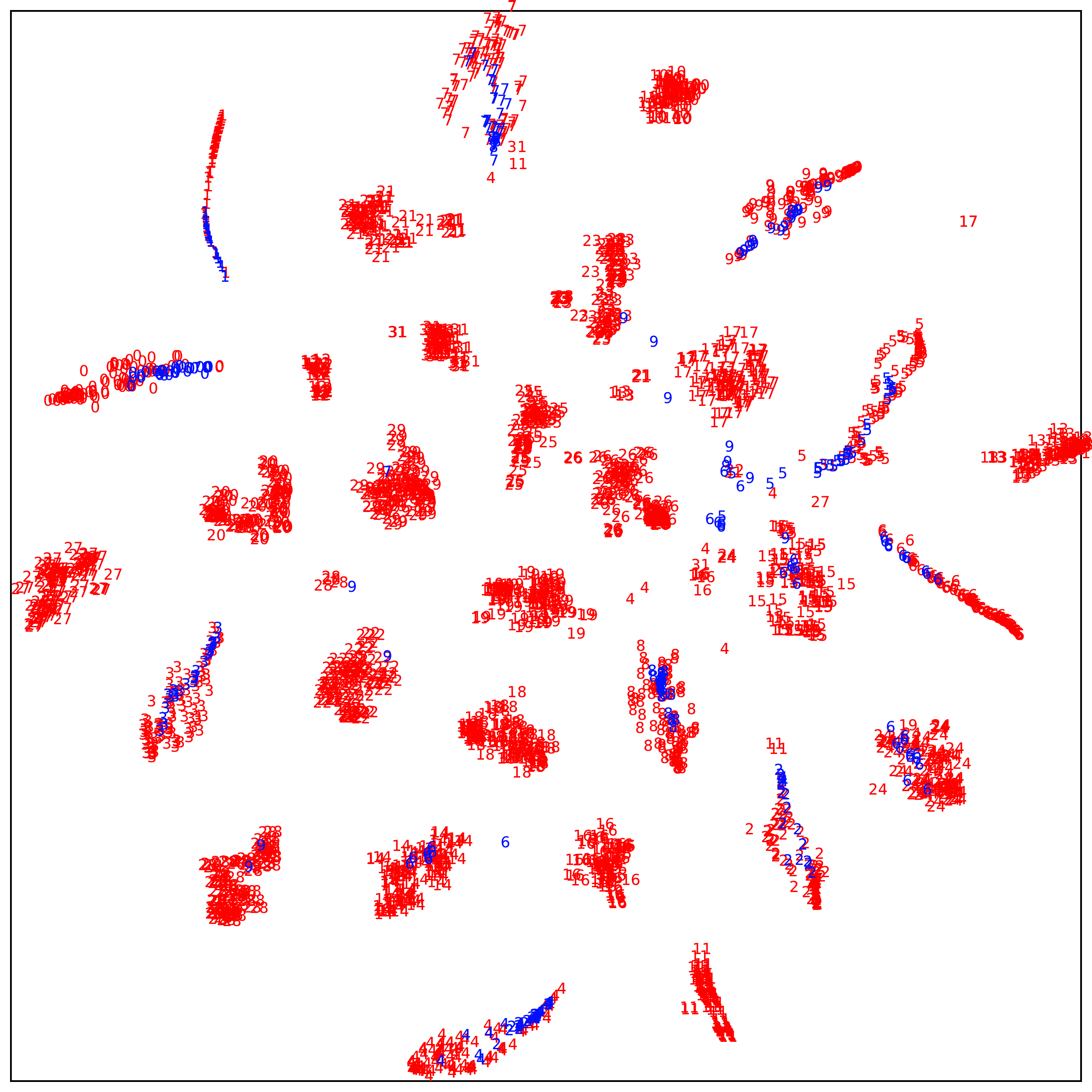}
  \caption{The t-SNE visualization of the alignment of source and target classes for the proposed method. The red numbers and blue numbers represent samples of source and target domain, respectively, and the values represent the classes.}\label{fig:labels}
\endminipage\hfill
\minipage{0.47\textwidth}
  \includegraphics[scale=0.31]{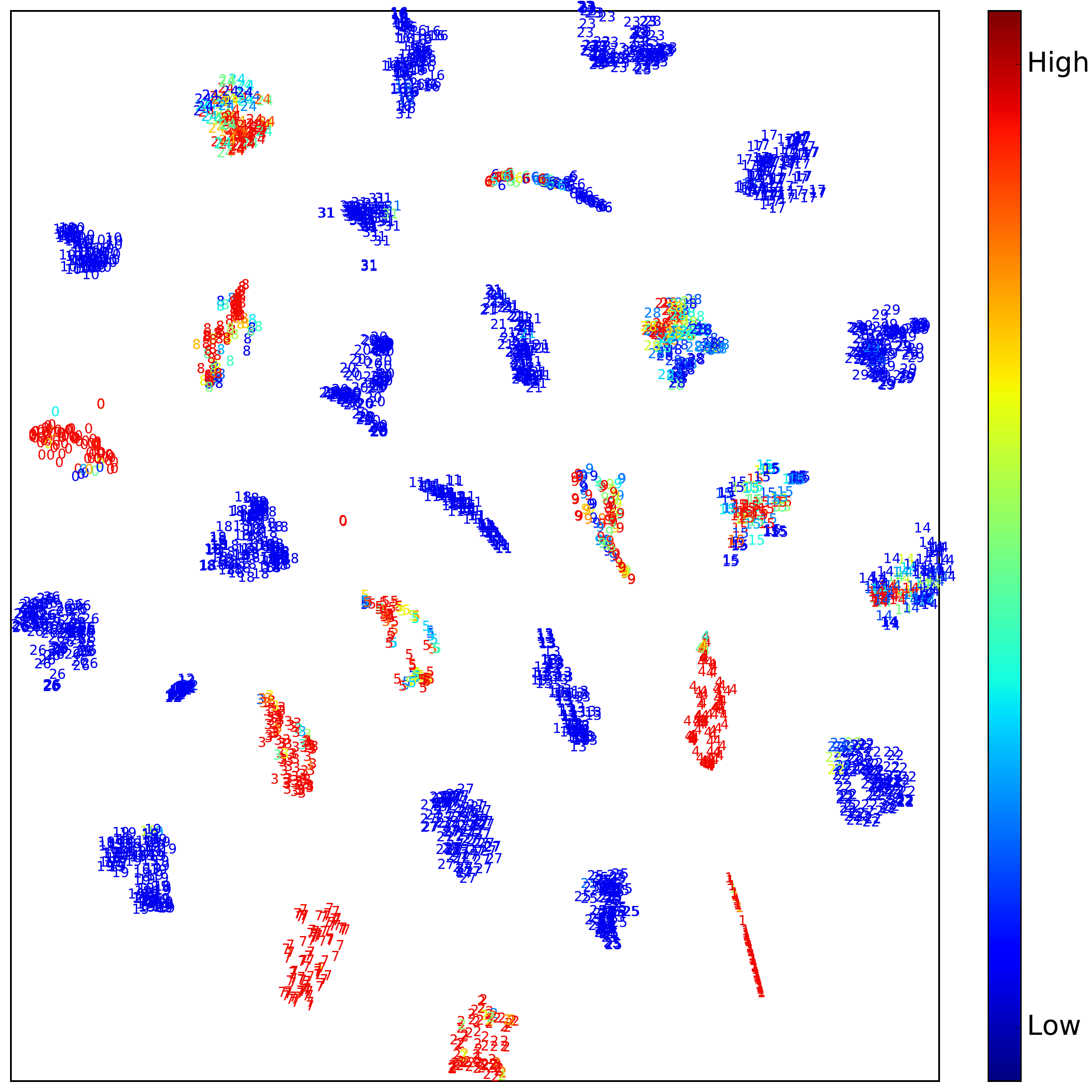}
  \caption{The t-SNE visualization of the learned weights of source samples for the proposed method. The red colored numbers indicate higher weights, while the blue colored numbers indicate lower weights. The intermediate values are arranged based
on the color bar.}\label{fig:weights}
\endminipage
\end{figure*}

The SAN methods have the same assumptions and perform comparably to the proposed method. However, a large number of domain classifiers are required in SAN compared to our method (i.e. the number of source classes v.s. two), which leads to far more parameters to train in SAN. The SAN-entropy is the SAN method without the target entropy minimization term, which corresponds to the proposed method with $\gamma=0$. The results in Table~\ref{tab:office} show that the proposed method ($\gamma=0$)  obtains better performance (86.73\%) on average than that of SAN (85.64\%), with much smaller number of parameters. 
\vspace{-1em}
\paragraph{Further analysis and evaluations}
For further verifying the effectiveness of the proposed weighting scheme, we also illustrate the alignment of the source and target class labels in Figure~\ref{fig:labels}. The same activations are used as in Figure~\ref{fig:bottleneck}. The ten classes in the target domain are labeled as 0$\sim$9 in blue which are the same set of classes as 0$\sim$9 in red in the source domain. Thus the number 10$\sim$30 in red are the outlier classes in the source domain. It shows that most of the target classes are aligned with the correct source classes. Figure~\ref{fig:weights} shows the learned weights using the first domain classifier $D_0$. If the weight of a source sample is large, the color of the sample tends to red while a smaller weight will be assigned with the blue color. The intermediate values are arranged based
on the color bar. It can be seen that most of the red coloured samples are from 0$\sim$9 classes while the outlier classes are mostly blue, demonstrating the effectiveness of the proposed weighting scheme for identifying samples from the outlier source classes.


\begin{table}[!ht]
\begin{small}
\begin{center}
\caption{Average accuracies (\%) on Caltech256$\rightarrow$Office10}
\vspace{-1em}
\begin{tabu}{|m{1cm}|m{0.6cm}m{1.3cm}m{1cm}m{0.8cm}m{1cm}|}
\hline
Methods & Alex & RevGrad\cite{Ganin2015} & RTN\cite{Long2016} & SAN\cite{Cao2017} & proposed \\
\hline
\hline
Average & 49.86 & 61.80 & 71.56 & \textbf{85.83} &  84.14 \\
\hline
\end{tabu}
\vspace{-1em}
\label{tab:cal}
\end{center}
\end{small}
\end{table}

\begin{table*}[!ht]
\begin{center}
\caption{The accuracy (\%) obtained on the Office-Caltech cross-domain object dataset}
\vspace{-1em}
\begin{tabular}{|m{2.7cm}|m{0.7cm}m{0.7cm}m{0.7cm}m{0.7cm}m{0.7cm}m{0.7cm}m{0.7cm}m{0.7cm}m{0.7cm}m{0.7cm}m{0.7cm}m{0.7cm}m{0.6cm}|}
\hline
Datasets & C10$\rightarrow$ A5 & C10$\rightarrow$ W5 & C10$\rightarrow$ D5 & A10$\rightarrow$ C5 & A10$\rightarrow$ W5 & A10$\rightarrow$ D5 & W10$\rightarrow$ C5 & W10$\rightarrow$ A5 & W10$\rightarrow$ D5 & D10$\rightarrow$ C5 & D10$\rightarrow$ A5 & D10$\rightarrow$ W5 & Avg. \\
\hline
\hline
AlexNet+bottleneck & 93.58 & 83.70 & 91.18 & 85.27 & 76.30 & 85.29 & 74.14 & 87.37 & \textbf{100.00} & 80.82 & 89.51 & \textbf{98.52} & 87.14\\
RevGrad~\cite{Ganin2015} & 91.86 & 82.22 & 83.82 & 77.57 & 65.93 & 80.88 & 72.60 & 80.30 & 95.59 & 69.35 & 77.09 & 80.74 & 79.83\\
RTN~\cite{Long2016} & 91.86 & 93.33 & 80.88 & 80.99 & 69.63 & 70.59 & 59.08 & 74.73 & \textbf{100.00} & 59.08 & 70.02 & 91.11 & 78.44\\
ADDA-grl~\cite{Tzeng2017} & 93.15 & 94.07 & 97.06 & 85.27 & \textbf{87.41} & \textbf{89.71} & 86.82 & 92.08 & \textbf{100.00} & 89.90 & 93.79 & \textbf{98.52} & 92.31\\
Proposed ($\gamma=0$) & 94.00 & \textbf{99.26} & 95.59 & \textbf{90.41} & \textbf{87.41} & 88.24 & 90.07 & \textbf{95.29} & \textbf{100.00} & 91.44 & \textbf{94.43} & \textbf{98.52} & 93.72\\
proposed & \textbf{94.22} & 97.78 & \textbf{98.53} & 89.90 & \textbf{87.41} & 88.24 & \textbf{90.24} & \textbf{95.29} & \textbf{100.00} & \textbf{91.61} & \textbf{94.43} & \textbf{98.52} & \textbf{93.85}\\
\hline
\end{tabular}
\label{tab:officeCal}
\end{center}
\end{table*}	
\vspace{-1em}
\begin{table*}[!ht]
\begin{center}
\caption{The accuracy (\%) obtained on the Office cross-domain object dataset}
\vspace{-1em}
\begin{tabular}{|l|lllllll|}
\hline
Datasets  & A31$\rightarrow$W10 & D31$\rightarrow$W10 & W31$\rightarrow$D10 & A31$\rightarrow$D10 & D31$\rightarrow$A10 & W31$\rightarrow$A10 & Avg. \\
\hline
\hline
AlexNet+bottleneck & 62.03 & 95.25 & 97.45 & 71.97 & 68.27 & 62.94 & 76.32 \\
RevGrad~\cite{Ganin2015} & 56.95 & 75.59 & 89.17 & 57.32 & 57.62 & 63.15 & 66.64 \\
RTN~\cite{Long2016} & 68.14 & 91.53 & 98.09 & 69.43 & 68.27 & 77.35 & 78.80 \\
ADDA-grl~\cite{Tzeng2017} & 63.39 & 98.31 & 98.73 & 73.25 & 70.46 & 72.34 & 79.41\\
SAN-selective~\cite{Cao2017} & 71.51 & 98.31 & \textbf{100.00} & 78.34 & 77.87 & 76.32 & 83.73\\
SAN-entropy~\cite{Cao2017} & 74.61 & 98.31 & \textbf{100.00} & 80.29 & 78.39 & 82.25 & 85.64\\
SAN~\cite{Cao2017} & \textbf{80.02} & 98.64 & \textbf{100.00} & \textbf{81.28} & 80.58 & \textbf{83.09} & 87.27\\
proposed ($\gamma=0$) & 75.25 & \textbf{98.98} & \textbf{100.00} & 80.25 & 84.66 & 81.21 & 86.73 \\
proposed & 76.27 & \textbf{98.98} & \textbf{100.00} & 78.98 & \textbf{89.46} & 81.73 & \textbf{87.57} \\
\hline
\end{tabular}
\label{tab:office}
\end{center}
\end{table*}

\begin{figure}[!h]
\includegraphics[scale=0.35]{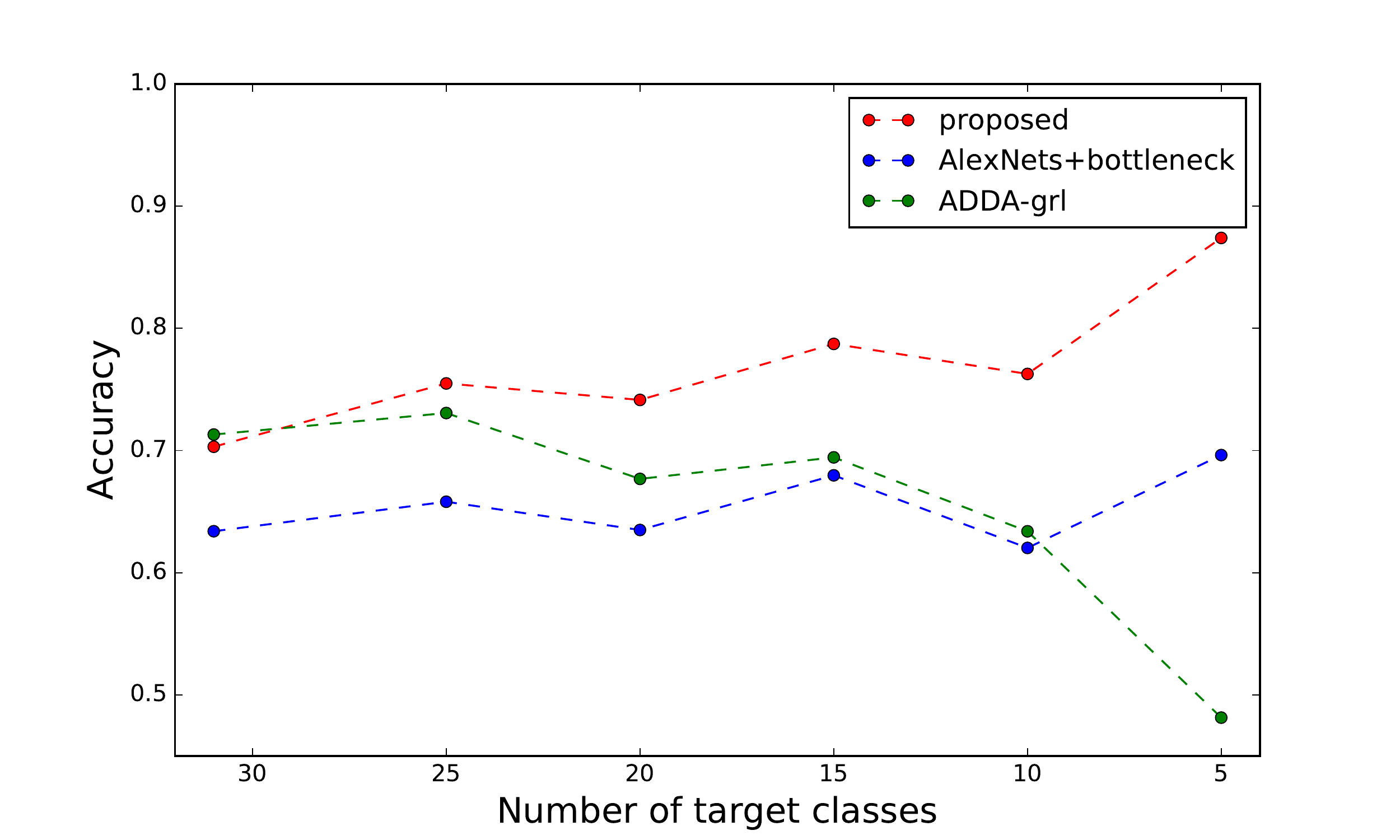}
\caption{The accuracy curve of varying the number of target classes for the baselines and the proposed method.}
\label{fig:Tclasses}
\end{figure}

We also conduct the experiments on evaluating the performance when the number of target domain classes varies. Figure~\ref{fig:Tclasses} shows the results on A$\rightarrow$W domain pair. The source domain has always 31 classes, but the number of target domain classes varies from 31 to 5, i.e. $\{31, 25, 20, 15, 10, 5\}$. The results show that the proposed method outperforms the AlexNets+bottleneck baseline largely in all cases. Specifically, when the number of target classes is getting smaller, the relative improvement is larger. It can also be observed that the less the target classes are, the lower the accuracy will be for the ADDA-grl method. Thus, when the number of target domain classes is unknown, our method can improve the performance consistently. 

To evaluate the proposed method on the traditional non-partial domain adaptation setting, we further conduct experiments on Office-31 and Office+Caltech-10 datasets using standard full protocol. The results in Table~\ref{tab:all} shows that no noticeable degradation is observed compared to the state-of-the-art methods.
\vspace{-1em}
\begin{table}[!h]
\begin{small}
\begin{center}
\caption{Average accuracies (\%) of non-partial setting}
\vspace{-1em}
\begin{tabu}{|m{1.5cm}|m{0.55cm}m{1.cm}m{0.55cm}m{1.1cm}m{1cm}|}
\hline
Methods & Alex & RevGrad & RTN & ADDAgrl & proposed \\
\hline
\hline
Office31 & 69.15 & 73.75 & 72.87 & \textbf{73.90} & 73.35 \\
OfficeCal10 & 86.10 & 90.90 & \textbf{93.40} & 92.21 & 91.71  \\
\hline
\end{tabu}
\label{tab:all}
\end{center}
\end{small}
\end{table}
\vspace{-1em}

To validate our statement that the unshared feature extractors can capture more domain specific features than a shared feature extractor, we compare the shared and unshared F networks on the most challenging domain pair A31$\rightarrow$W10, and the results are 71.5\% for shared, and 76.3\% for unshared.
\vspace{-0.5em}
\section{Conclusion}
This paper extends the adversarial nets-based unsupervised domain adaptation to partial domain adaptation. A weighting scheme based on the activations of the adversarial nets is proposed for detecting the samples from the source domain outlier classes to effectively reduce the shift between the target data and the source data that are within the target classes. The experimental results show that the proposed method outperforms previous domain adaptation methods to a large degree and is comparable to the state-of-the-art partial transfer methods, demonstrating the effectiveness of the proposed method. For the future work, we will further exploit the method with the focus on larger scale partial domain adaptation.
\newpage


\newpage
\FloatBarrier

\begin{small}
\bibliographystyle{ieee}
\bibliography{CrossDataset}
\end{small}

\end{document}